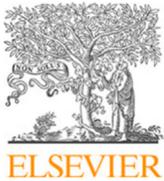

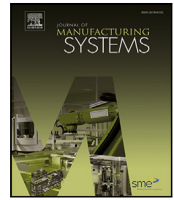

Technical paper

# Real-time decision-making for Digital Twin in additive manufacturing with Model Predictive Control using time-series deep neural networks

Yi-Ping Chen [a], Vispi Karkaria [a], Ying-Kuan Tsai [a], Faith Rolark [a], Daniel Quispe [a], Robert X. Gao [b], Jian Cao [a], Wei Chen [a],*

[a] Department of Mechanical Engineering, Northwestern University, United States of America
[b] Department of Mechanical and Aerospace Engineering, Cast Western Reserve University, United States of America

## ARTICLE INFO

## ABSTRACT

Digital Twin – a virtual replica of a physical system enabling real-time monitoring, model updating, prediction, and decision-making – combined with recent advances in machine learning, offers new opportunities for proactive control strategies in autonomous manufacturing. However, achieving real-time decision-making with Digital Twins requires efficient optimization driven by accurate predictions of highly nonlinear manufacturing systems. This paper presents a simultaneous multi-step Model Predictive Control (MPC) framework for real-time decision-making, using a multivariate deep neural network, named Time-Series Dense Encoder (TiDE), as the surrogate model. Unlike conventional MPC models which only provide one-step ahead prediction, TiDE is capable of predicting future states within the prediction horizon in one shot (multi-step), significantly accelerating the MPC. Using Directed Energy Deposition (DED) additive manufacturing as a case study, we demonstrate the effectiveness of the proposed MPC in achieving melt pool temperature tracking to ensure part quality, while reducing porosity defects by regulating laser power to maintain melt pool depth constraints. In this work, we first show that TiDE is capable of accurately predicting melt pool temperature and depth. Second, we demonstrate that the proposed MPC achieves precise temperature tracking while satisfying melt pool depth constraints within a targeted dilution range (10%–30%), reducing potential porosity defects. Compared to Proportional–Integral–Derivative (PID) controller, the MPC results in smoother and less fluctuating laser power profiles with competitive or superior melt pool temperature control performance. This demonstrates the MPC's proactive control capabilities, leveraging time-series prediction and real-time optimization, positioning it as a powerful tool for future Digital Twin applications and real-time process optimization in manufacturing.

## 1. Introduction

Autonomous manufacturing is essential for achieving efficiency, precision, and adaptability in modern production, leading to faster, reliable processes through machine learning integration [1,2]. Digital Twins, an emerging paradigm for manufacturing, serve as virtual counterparts of physical systems, facilitating bi-directional interaction, prediction, and decision-making under varying operational conditions and uncertainties where the digital and the physical systems evolve together [3–5]. In manufacturing, Digital Twins enable proactive real-time decision-making to optimize and control the process under rapid changes in operation and uncertain conditions [6,7]. One key element of realizing the Digital Twins in manufacturing is to perform real-time decision-making using the virtual model, providing proactive control actions to ensure part quality. Take additive manufacturing (AM) as an example, proactive control approaches, by incorporating process constraints and real-time process feedback, empower AM to anticipate and prevent deviations that could lead to defects [8,9], thereby improving part quality and streamlining the manufacturing process. However, constructing a real-time decision-making framework in AM is challenging due to the complexity and computational requirement of the decision-making process and the need for a compatible physics-based model [10].

Conventional feedback control methods, such as proportional–integral–derivative (PID) controllers, offer fast and reactive control actions and have been widely applied in AM [11], but struggle with the dynamic, multi-variate, and stochastic nature of AM processes. PID control schemes are typically SISO (single-input single-output) systems, meaning they adjust only a single process parameter, such as laser power, to achieve a single control objective, like maintaining a specific melt pool size or temperature. This approach overlooks the complex

**Nomenclature**

| | |
|---|---|
| $\hat{f}(\cdot)$ | Surrogate model of the system |
| $\hat{x}_k$ | Predicted state at timestep $k$ |
| $d_x, d_y$ | Distances from the current laser position to the nearest geometry boundary along the $x-$ and $y-$axis |
| $J(\cdot)$ | Objective function for MPC |
| $p$ | Horizon length |
| $u_k$ | Control input at timestep $k$ |
| $w$ | Window size |
| $x_k$ | Actual state at timestep $k$ |
| $z$ | $z-$coordinate of the laser nozzle position |
| AM | Additive Manufacturing |
| DED | Directed Energy Deposition |
| DNN | Deep Neural Network |
| MPC | Model Predictive Control |
| NN | Neural Network |
| TiDE | Time Series Dense Encoder |

interactions between multiple process parameters and their combined effects on the final product [1], limiting their applicability in AM. Also, PID-based control is considered reactive, as it responds to errors only after they occur rather than anticipating and preventing them. For example, in [12], experimental results show that PID control fails to maintain melt pool width stability, leading to geometric inaccuracies. This is largely because PID controllers use fixed parameters throughout the build process, whereas AM conditions change over time due to heat accumulation. As a result, a single optimal set of controller parameters cannot be maintained for the entire process. In addition, they do not inherently account for constraints (e.g., physical limits on actuators or process variables), especially when multiple parameters or changing conditions are involved [1].

In contrast, model predictive control (MPC) is gaining popularity for its ability to predict future behaviors, optimize control inputs for multi-input multi-output (MIMO) systems, and handle constraints [13–15], which is a crucial feature for proactive defect mitigation [1,16]. For example, constraints in AM include maintaining a specific melt pool temperature while ensuring the melt pool depth remains within a defined range. The MPC optimization framework, with its capability of state prediction and constraints handling in real-time decision-making processes, provides more robust and reliable control solutions [17,18] and enables adaptive decision-making in Digital Twins [1,6].

Although MPC has been extensively studied in process control, its application in AM for controlling melt pool features is still limited. Specifically, significant gaps remain in the demonstration of effective constraint handling for quality control that builds on the accurate nonlinear surrogate modeling of the AM system while supporting the solving speed/latency of the MPC. Most existing approaches use linear state-space models and transfer functions derived from system identification (ID) methods [19–21]. While the analytical solutions for optimal control problems can be obtained, e.g., using linear quadratic regulator (LQR), they often struggle to capture AM's highly nonlinear dynamics without considering constraints. This restricts MPC in AM to operate under limited conditions, e.g., the model is only accurate in one layer [21]. To address this limitation, one solution is to use multiple linear models to approximate different behaviors under varying conditions [21]. While experimentally validated, these approaches demonstrate control capabilities mainly under steady-state conditions due to the high variability of melt pool dynamics. Aside from system ID methods, Cao et al. proposed a simplified semi-ellipse approximation for melt pool control, using MIMO robust MPC to adjust laser power and scanning speed [22]. Although this model accounts for more physics, its first-order approximation may not fully represent the complexities of heat distribution and dissipation. Also, similar to the PID controllers, the linear model representations limited their adaptability to changing conditions where the system dynamic may vary drastically. In fact, embedding all relevant physics into a real-time model is nearly impossible due to unobservable states and parameter estimation challenges. Given the success of machine learning (ML), data-driven methods offer a promising alternative for AM, capturing nonlinear dynamics, and enabling the MPC to mitigate defects proactively.

The rise of ML and deep neural networks (DNN) has fueled interest in data-driven approaches for MPC, allowing for nonlinear system identification and robust control [23]. These models offer flexibility and require fewer assumptions, making them useful across many engineering applications [24–28]. However, most applications use state-space models to represent the dynamic controlled systems. State-space models, even with nonlinear identification [29], need the assumption of fully observable systems, which presents a challenge in AM in practice. In addition, they often oversimplify the dynamics by treating systems as Markov processes whose future state of a system depends only on its present state and not on its past states. Autoregressive models, by including a longer history of the past states, capture complex, long-term dependencies and provide additional information as compensation for partially observable systems [30–32]. This feature is better suited for complex systems like AM [32], as they learn both dynamics and predict states more effectively. On the other hand, unlike one-step predictive models that require costly recursive rollouts for the MPC, simultaneous multi-step MPC, first proposed by Park et al. [33], with sequence output, provides one-shot predictions for future states simultaneously across the entire horizon, significantly improving both speed and accuracy. Thus, in this work, we aim to explore the full potential of integrating time-series models with AM, not only to learn the complex system dynamics, but also to develop an efficient MPC framework that mitigates defects proactively.

In this study, we present a simultaneous multi-step MPC framework using a time-series DNN as a proof of concept of real-time decision-making for manufacturing systems, using Directed Energy Deposition (DED) AM as our case study. However, we envision that this method can be generalized for similar manufacturing processes such as welding. The framework dynamically regulates laser power to maintain the target melt pool temperature while satisfying constraints on melt pool depth. As shown in Fig. 1, a multi-variate time-series model, named Time Series Dense Encoder (TiDE) [34], is first trained offline to predict melt pool temperature and depth, serving as the surrogate model for the MPC. During the online MPC, a constrained optimization problem is solved to compute the optimal sequence of control inputs that minimize the temperature tracking error while satisfying depth constraints. The first element of this sequence is applied to the physical system, where the powder deposition and melting process are represented by simulation based on a finite element analysis (FEA) code in this work, to execute the control action. Then, the resulting melt pool temperature and depth are extracted and fed back into the MPC, completing the closed-loop control process.

The major contributions of this work include:

- We propose a multi-step MPC framework for realizing the real-time decision-making of the Digital Twin concept that leverages a time-series-based deep learning model using the AM system as an example. The multi-step output model structure along with automatic differentiation, accelerates the MPC and enables fast solutions for real-time optimization.
- We demonstrate the use of a multi-variate physics-based time-series deep learning model for predicting critical local behaviors in manufacturing systems, and it is capable of accurately forecasting states across the prediction horizon in a one-shot manner.

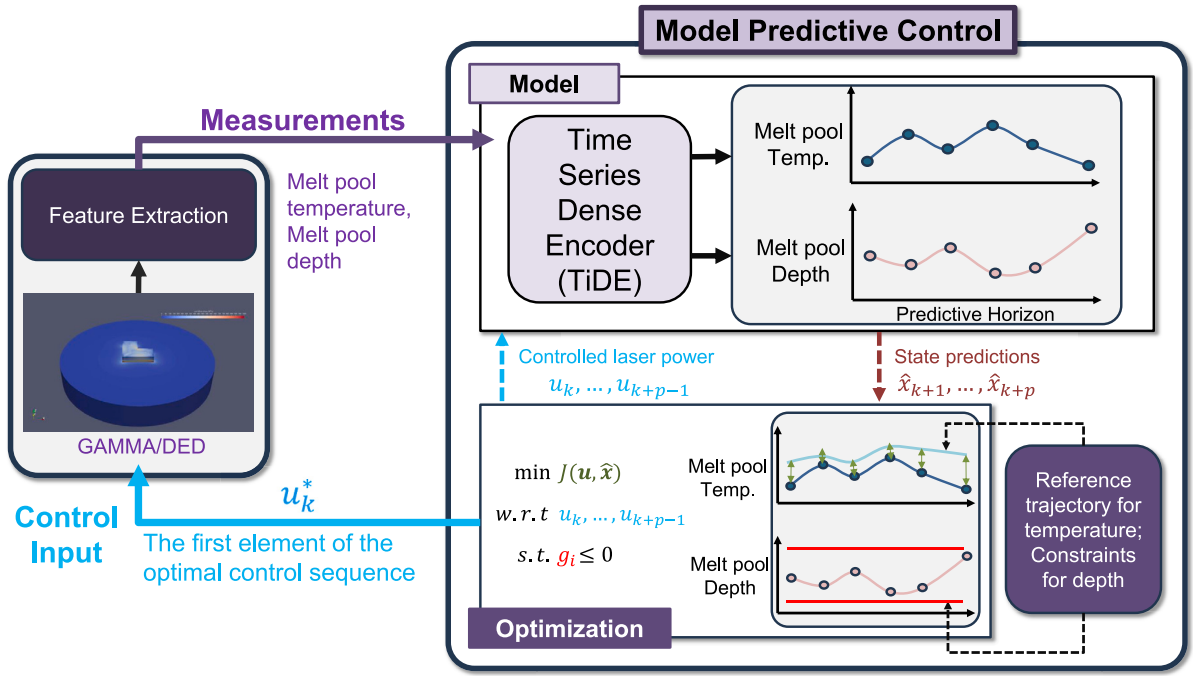

**Fig. 1.** System overview of the proposed multi-step MPC framework for DED, where $J$ is the cost function, *w.r.t* denotes "with respect to", *s.t.* denotes "subject to", and $g_i$ represents the $i$th constraint.

- We demonstrate effective constraint handling directly on state observations in DED, providing a proactive approach to mitigate defects.

The remainder of the paper is structured as follows: Section 2 provides an overview of the technical background on time-series modeling and the MPC, while Section 3 covers the thermal simulation model. In Section 4, we outline the data preparation process for model training, followed by the development of the multi-step MPC in Section 5. Section 6 presents and discusses the results and comparisons, and we conclude the paper in Section 7.

## 2. Technical background

### 2.1. Model predictive control

Model Predictive Control (MPC), also known as receding horizon control, is an advanced control technique that uses a model of the system to predict its future behavior and optimizes control actions by solving a finite-horizon optimal control problem at each sampling instant [13], as illustrated in Fig. 2. In standard MPC, once the control sequence is solved at the current step, the first action is applied to the plant, and the process is repeated as the system advances to the next step with updated observations. Assuming the prediction horizon is $p$, the MPC to optimize future control inputs $\mathbf{u}$ given the current state $\mathbf{x}_k$ at current time $k$ can be formulated as:

$$\min_{\mathbf{u}=[\mathbf{u}_k,...,\mathbf{u}_{k+p-1}]} J(\mathbf{u}, \mathbf{x}_k) = \sum_{i=0}^{p-1} \left[ \|\mathbf{x}_{k+i}\|_{\mathbf{Q}}^2 + \|\mathbf{u}_{k+i}\|_{\mathbf{R}}^2 \right], \quad (1a)$$

$$s.t. \quad \hat{\mathbf{x}}_{k+i+1} = \hat{f}(\hat{\mathbf{x}}_{k+i}, \mathbf{u}_{k+i}), \ \forall i \in \mathbb{N}_{[0,p]}, \quad (1b)$$

$$\mathbf{x}_{k+i} \in \mathbb{X}, \ \forall i \in \mathbb{N}_{[0,p-1]}, \quad (1c)$$

$$\mathbf{u}_{k+i} \in \mathbb{U}, \ \forall i \in \mathbb{N}_{[0,p-1]}, \quad (1d)$$

where $\|\mathbf{x}\|_{\mathbf{Q}}^2 = \mathbf{x}^\top \mathbf{Q}\mathbf{x}$ represents the quadratic operation of state vector $\mathbf{x}$, the weighting matrices $\mathbf{Q} > 0$ and $\mathbf{R} > 0$ are symmetric. Eq. (1b) is the general representation of the dynamic equation in which $\hat{f}$ is the predictive model, and Eqs. (1c) and (1d) are the constraints of states and control actions, respectively.

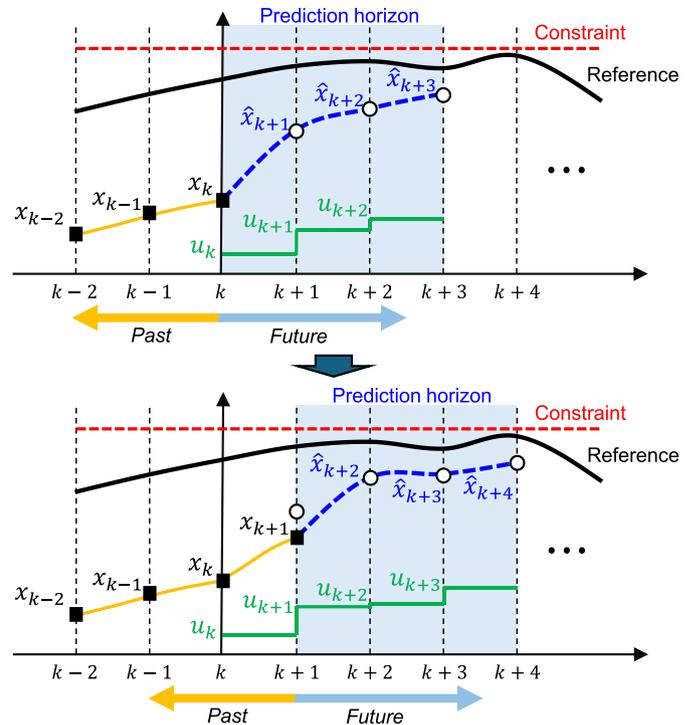

**Fig. 2.** Illustration of MPC with the moving horizons step-by-step.

### 2.2. Time series model: Time Series Dense Encoder (TiDE)

When selecting an appropriate predictive model for the MPC using the time series DNN, two key considerations are essential: (1) The inference speed as MPC requires multiple function evaluation in each iteration. (2) The data format that is compatible with the DED. In this work, we selected Time Series Dense Encoder (TiDE) [34], a residual neural network (ResNet) specifically designed for multi-variate time series forecasting, due to its efficiency and accuracy, as shown in Fig. 3. Its residual connection feature allows it to capture long-term dependencies of the history without vanishing gradient. Moreover, it utilizes a dense coder to extract key features from input data, acting as a filter to enhance the resilience of the model against disturbances in state feedback. TiDE operates in linear time complexity, making it faster than other DNN-based time-series models like RNNs, which require recursive rollouts for sequence output, and Transformer-based methods, e.g. PatchTST [35], which involve a minimum $O(n^2)$ complexity due to self-attention mechanisms [34]. Moreover, TiDE exhibits strong predictive performance in several benchmarks [34] and can quantify aleatoric (data) uncertainty in a one-shot manner using Gaussian or quantile regression [36]. Note that different from our previous work [7] where a Bayesian LSTM is proposed to predict the complete temperature profile for the whole part for offline optimization, in this work, TiDE model is an accurate representation of local system behavior, which is more suitable for the MPC.

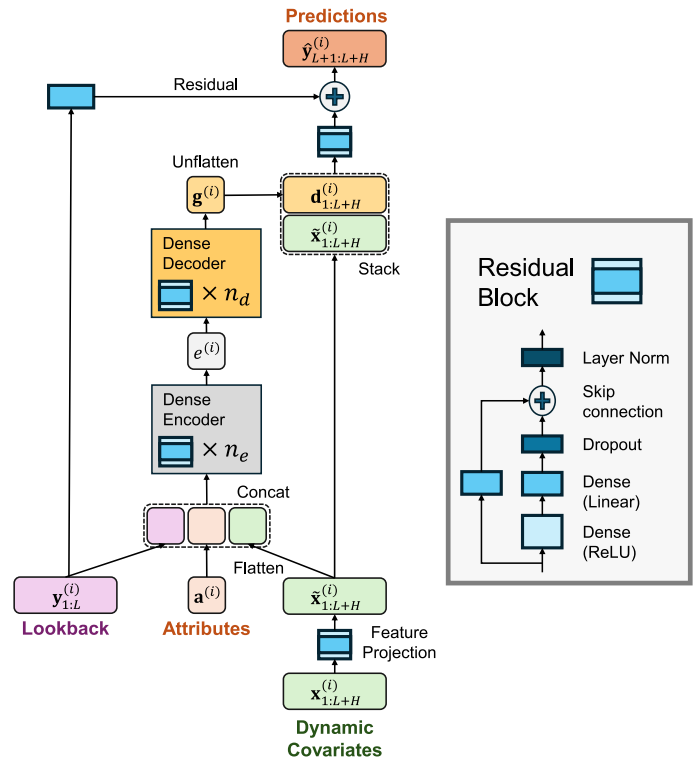

**Fig. 3.** Network structure of TiDE.
*Source:* Modified from [34].

The data format that TiDE requires also matches our system as its capability of handling covariates is a critical feature of TiDE in surrogating dynamical systems and MPC applications, which cannot be seen from either the conventional Transformer Encoder or PatchTST. This feature is crucial for surrogating dynamical control systems because the prediction output *should* condition on the future inputs. For TiDE, not only the past states but also the future control inputs and the geometry variables, as a form of dynamic covariates, are treated as predictive inputs to provide rich information on forecasting the future states (target) in a one-shot manner. This is a significant advantage over some sequence-to-sequence models that do not support future covariates, (e.g. Transformer encoders, Long Short-Term Memory (LSTM), N-BEATS [37]) where the user has to mask out future responses to structure the data into a compatible format [33] and not fulfilling the whole potential of the model. The structure of the model input, predictive target, and the role of geometric information as covariates will be elaborated in Section 4.4.

In addition, TiDE's embedding structure, realized through dense encoders and decoders, enhances predictions by transforming raw inputs into dense, low-dimensional representations that capture meaningful patterns and relationships. The encoder reduces data dimensionality, encodes complex interactions, and improves the model's ability to generalize across unseen examples. The dimensionality-reduced embeddings also function as noise filters by identifying and embedding only the most relevant features in the latent space. By effectively compressing input information, these embeddings mitigate the risk of overfitting, particularly in high-dimensional datasets. For time-series data, embeddings efficiently represent temporal attributes or categorical features, allowing TiDE to extract richer patterns and improve prediction accuracy.

## 3. Thermal simulation model

### 3.1. Explicit finite element solver

In this work, an in-house developed explicit FEA code is employed for part-scale transient heat transfer simulations of the DED process [38]. The code, named GAMMA, is implemented in Python with temperature-dependent material properties and accelerated by GPU computation using CuPy 9.0.0 [39]. The heat transfer model is based on Fourier's law of heat conduction, with the governing equation expressed as:

$$\rho C_p(T)\frac{\partial T}{\partial t} = k(T)\nabla^2 T + q_{laser} + q_{conv} + q_{rad}, \tag{2}$$

where $T$ is the temperature (K), $\rho$ is the material density (g/mm$^3$), $C_p(T)$ and $k(T)$ are the temperature-dependent specific heat capacity (J/g/K) and thermal conductivity (W/m/K), respectively. $q_{laser}$ is the heat flux (W/m$^2$) from the laser, and $q_{conv}$ and $q_{rad}$ represent the convective and radiative heat fluxes (W/m$^2$). This simulation applies heat flux boundary conditions, including a laser surface flux boundary condition $q_{laser}$ to represent heat provided from a Gaussian laser beam applied to only the top surface of elements, since the laser is projected from the top-down, expressed as:

$$q_{laser} = \frac{-2\eta P}{\pi r_{beam}^2} \exp\left(\frac{-2d^2}{r_{beam}^2}\right), \tag{3}$$

where $\eta$ is the absorption coefficient (%), $P$ is the laser power (W), $r_{beam}$ is the beam radius (mm), and $d$ is the distance (mm) from the node to the center of the laser. Additionally, boundary conditions for convective heat loss $q_{conv}$ and radiative heat loss $q_{rad}$ are applied to all exposed surfaces of the part excluding the bottom substrate surface:

$$q_{conv} = h(T - T_0), \tag{4}$$

$$q_{rad} = \sigma\epsilon(T^4 - T_0^4), \tag{5}$$

where $h$ is the convection heat transfer coefficient (W/m$^2$/K), $\sigma$ is the Stefan–Boltzmann constant ($5.67 \times 10^{-8}$ W/m$^2$/K$^4$), $\epsilon$ is the material's emissivity, and $T_0$ is the ambient temperature. The bottom surface of the substrate employs a Homogeneous Dirichlet boundary condition to fix displacement and maintain an isothermal temperature at 300 K (room temperature):

$$T|_{z=0} = T_0. \tag{6}$$

Material deposition is modeled using the inactive element method. In the preprocessing stage, the activation time of each element is determined based on the predefined toolpath. An element becomes active and is incorporated into the mesh when the distance between the laser center and the geometric center of the element is smaller than the beam diameter $r_{beam}$. One can refer to our previous work for more details about the FEA model setup [38,40].

### 3.2. Calibration approach for thermal simulation model

Since GAMMA simulates only heat conduction for part-scale simulations and neglects Marangoni flow, which contributes to convective heat dissipation, it overestimates melt pool temperatures compared to sensor values—particularly when the temperature exceeds the material's liquidus point. One way to compensate for this is by calibrating the material's artificial thermal conductivity using experimental data when an element's temperature surpasses the liquidus threshold [40,41], and the effectiveness of this approach has been validated using Inconel 718 in our previous work [40]. For simplicity, in this work, we multiplied the extracted melt pool temperature by 0.5 to emulate a reasonable temperature value, compensating for the overestimation of melt pool temperature. This treatment will be refined in our future work for 316L. Since the current research scope does not involve real physical data as feedback for model predictive control but directly uses the GAMMA simulation result to emulate the true physical system, the simplistic scaling approach will not affect the demonstration of TiDE and MPC's learning and control capabilities or undermine our method's effectiveness.

## 4. Data and model preparation

### 4.1. Target geometry and material

In this work, we chose a single-track square as the target geometry made of 316L on a thick substrate of AISI 1018, as shown in Fig. 4 and the specifications in Table 1. This part allows the MPC to run for a long distance in a straight line before sharp turning and switching layers, and the single-track wall on each side allows the verification of melt pool depth using the infrared camera. The geometry and mesh were generated in ABAQUS CAE 2023. The mesh was comprised of linear hexahedral elements (0.375 mm in size). The substrate was partitioned to coarsen the mesh outside of the deposition area to reduce the number of elements and decrease computational time. A corresponding FEA model is built and simulated using GAMMA, and the temperature profiles on each node are saved for feature extractions and model training.

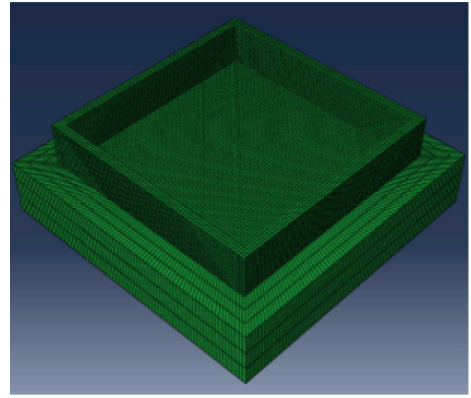

**Fig. 4.** Single-track square.

**Table 1**
Specification of printed square.

| Item | Quantity |
| --- | --- |
| Side length | 40 mm |
| Track width | 1.5 mm |
| Layer height | 0.75 mm |
| Num. of layers | 10 layers |
| Element size | 0.375 mm |
| Num. of elements | 40 540 |
| Substrate height | 10 mm |
| Scanning speed | 7 mm/s |

### 4.2. Feature extraction

We developed an algorithm to extract the target features – melt pool temperature and depth – from the temperature profiles of all the nodes that belong to the geometry. These two features are considered the key to maintaining printing quality and mitigating defects (elaborated in Section 5). This algorithm runs one sampling timestep every five GAMMA simulation time steps (1 sampling time step = $5 \times 0.00714$ s/step) to save memory sizes. As the laser location at each sampling time step is known, the melt pool temperature can be extracted by the following steps, also illustrated in Fig. 5(a):

1. Select the activated nodes that belong to the top layer of the current printed geometry (current max laser location).
2. Among the selected nodes, further select the nodes around the laser location, i.e., ±3 mm on $x$- and $y$- directions, centered on the laser location.
3. Fit a radial basis function (RBF) surface [42] using the nodes selected from the previous step since the selected nodes are sparsely distributed. Then, the RBF surface will be interpolated with finer mesh grids (0.2 mm) to get a higher resolution of the temperature map, reducing the numerical errors induced by the coarse meshes.
4. Calculate the average temperature within the scanning radius (0.9 mm) around the laser location as the melt pool temperature $x_{temp}$. This approach emulates how coaxial photodiodes measure temperature in a physical DED machine as it only calculates the mean temperature within its sensing region.

We also developed an algorithm for extracting melt pool depth from the simulation. Note that although the in-situ measurement of melt pool depth may not be available for most DED machines, it can still be obtained via online estimation methods [43–46]. Here, we assume that the melt pool depth is accessible in-situ. The algorithm can be detailed by the following steps, as illustrated in Fig. 5(b):

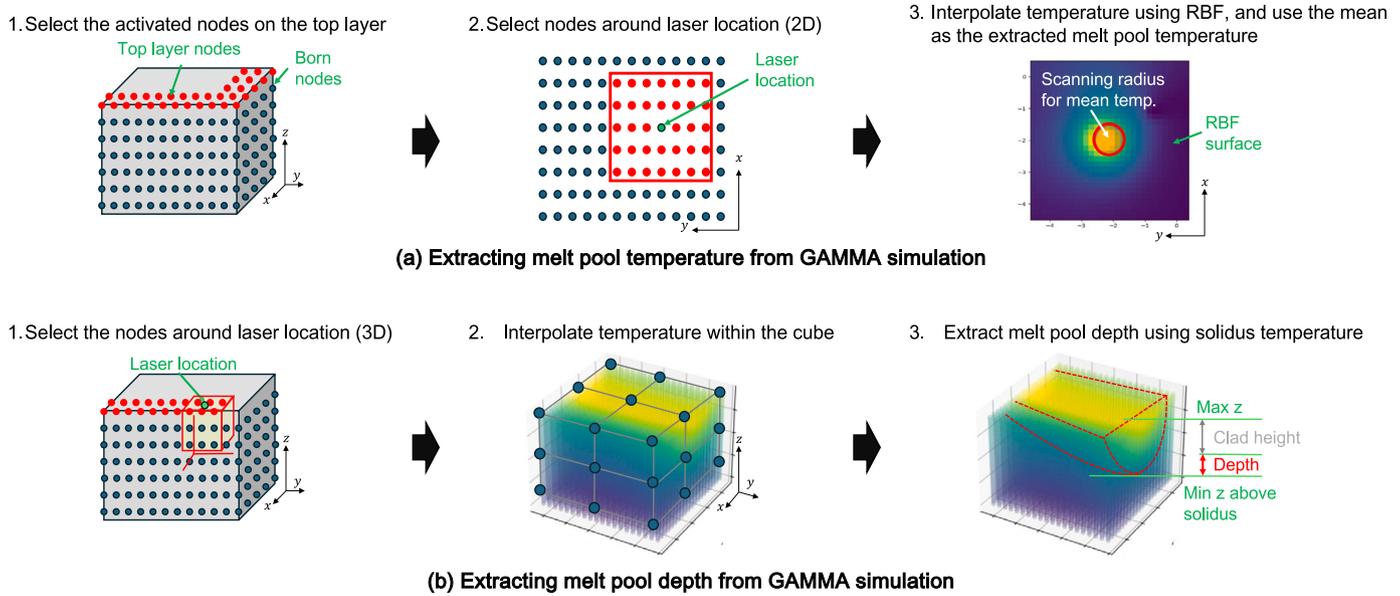

1. Select the activated nodes on the top layer
2. Select nodes around laser location (2D)
3. Interpolate temperature using RBF, and use the mean as the extracted melt pool temperature

**(a) Extracting melt pool temperature from GAMMA simulation**

1. Select the nodes around laser location (3D)
2. Interpolate temperature within the cube
3. Extract melt pool depth using solidus temperature

**(b) Extracting melt pool depth from GAMMA simulation**

**Fig. 5.** Feature extraction processes.

1. Select nodes within a cube of size 1.5×1.5×4 mm centered on the laser, extending ±0.75 mm on $x$- and $y$- directions, and −4 mm in $z$- direction.
2. Fit a three-dimensional RBF using the selected nodes from the previous step, with its coordinates as input and corresponding temperature as output. Interpolate the RBF with finer grid meshes.
3. Extract the melt pool depth by calculating the maximum distance along the $z$-axis for all grid points where the temperature exceeds the solidus temperature of 316L. Subtract the clad/layer height (0.75 mm in this case) to obtain the final melt pool depth, $x_{depth}$.

Note that the melt pool depth obtained using this algorithm can be negative due to the heat conduction simulation, especially in the first three layers and at the start of a new layer. In this work, we will keep the raw data as it is collected without modifying/clamping the negative melt pool depth extractions to reflect the GAMMA simulation even though it is different from the experimental setting. The feature extraction algorithm will be used in both data collection offline, and to emulate the function of the in-situ sensors in the online MPC.

### 4.3. Data collection

The objective of data generation is to create a series of laser power profiles that uniformly span the design space throughout the entire printing process. Given the high dimensionality of the laser power profile design space, traditional design of experiments (DOE) methods are impractical. To address this issue, we adopted the approach proposed by Karkaria et al. [7], which represents each laser power profile using 10 key parameters. These parameters include the amplitude, number of terms, frequency, and phase from the Fourier series approximation, as well as the rate of change for the amplitude, frequency, and phase of the wave. Additionally, three parameters account for the slope, fluctuation, and amplitude of the seasonal component of the laser power time series, providing greater flexibility in the representation. This dimensional reduction enables the application of DOE with 10 parameters using the optimal Latin hypercube sampling method [47]. These designs can subsequently be used to reconstruct temperature profiles for the entire print process. In this work, 100 laser power profiles are generated using this method.

As the laser power profiles are generated, part-scale simulations of the printing process are performed using GAMMA, with the generated laser profiles serving as input under ideal, noiseless, open-loop conditions. For convenience, all 100 time series are concatenated into a single continuous time series. Key features of interest, such as melt pool temperature ($x_{temp}$) and melt pool depth ($x_{depth}$), are extracted using the aforementioned algorithm and saved as time series profiles upon the completion of the simulation. To mitigate the effect of the numerical errors introduced during the GAMMA simulation, both melt pool temperature and depth are smoothed using the moving average method with a window size of four. In addition to these features and the laser input ($u$), three other parameters are recorded at each time step: the $z$-coordinate of the laser position, and the distances from the current laser position to the nearest geometry boundary along the $x$-axis and $y$-axis, denoted as $d_x$ and $d_y$, respectively. Each time series profile is further divided into snapshots using a moving window approach with a step size equal to one. Consequently, each segment has a length of $w+p$, where $w$ represents the window size (i.e., the length of the history) and $p$ represents the prediction horizon. The resulting data collected from this stage are represented as $\mathbf{x}_{temp}^i \in \mathbb{R}^{w+p}, \mathbf{d}_x^i \in \mathbb{R}^{w+p}, \mathbf{z}^i \in \mathbb{R}^{w+p}, \mathbf{x}_{depth}^i \in \mathbb{R}^{w+p}, \mathbf{d}_y^i \in \mathbb{R}^{w+p}, \mathbf{u}^i \in \mathbb{R}^{w+p}, \forall i \in \mathbb{N}_{[1,N]}$, where $N$ is the total number of fractions of time series, as visualized in Fig. 6.

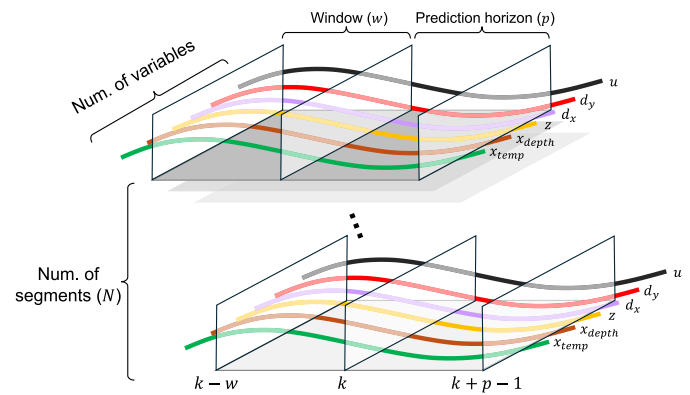

**Fig. 6.** Segmenting training data with moving window.

### 4.4. Model training

TiDE is a forecasting model that supports data in a specific time-series format [34]. In forecasting, a *covariate* is an external variable that influences the *target*, the quantity to be predicted, but is not the target itself, providing context for the predictive model [48]. Past covariates are obtained in previous time steps, while future covariates are known or estimated inputs affecting future target predictions. This framework aligns well with dynamical systems and multi-step MPC, where control inputs serve as covariates and observations or states represent the target.

In this work, we defined melt pool temperature ($x_{temp}$, in K) and depth ($x_{depth}$, in mm) as the target, and the distances to part boundaries in $x$ and $y$ ($d_x, d_y$, in mm), laser $z$ position ($z$, in mm), and laser power ($u$, in W) as covariates, as shown in Fig. 7. The relationship between covariates and target is represented by:

$$\begin{bmatrix} \hat{x}_{temp,k+1:k+p} \\ \hat{x}_{depth,k+1:k+p} \end{bmatrix} = \mathrm{TiDE}(\mathbf{x}_{temp,k-w+1:k}, \mathbf{x}_{depth,k-w+1:k}, \mathbf{d}_{x,k-w:k+p-1},$$
$$\mathbf{d}_{y,k-w:k+p-1}, \mathbf{z}_{k-w:k+p-1}, \mathbf{u}_{k-w:k+p-1}). \tag{7}$$

A key advantage of TiDE is its ability to directly incorporate both past and future covariates, along with past targets, to predict future targets. Unlike most generic sequence-to-sequence models, such as RNNs, GRUs, LSTMs, and Transformers, which do not distinguish between covariates and targets and require target masking to align with the input format [33], TiDE maintains full predictive power without leaving any network parameters idle. While static covariates, shown in Fig. 7, are not used in this study, they offer additional flexibility for future work, such as encoding different material properties or part geometry.

We chose quantile loss [49] as the loss function for our model training because it offers several advantages. First, quantile loss is less sensitive to outliers than mean square error (MSE) loss, making it more robust when dealing with noisy or skewed data. Second, for probabilistic learning, in contrast to Gaussian loss [50] which learns the mean and variance during training, it does not require any assumptions about the underlying data distribution to learn and predict specific quantile levels. Moreover, with quantile prediction, it can directly quantify aleatoric (data) uncertainty without the need for Monte Carlo sampling or Bayesian inference. Although in this work we do not consider the uncertainty quantification of the model, the fast prediction of data uncertainty makes it well-suited for future applications toward robust MPC. In the rest of the work, we used the predicted median (0.5 quantile) as the response prediction. The PyTorch realization for both TiDE and quantile loss are modified from [36] to enable automatic differentiation.

The general idea of selecting the horizon length is that it should be longer than the principle dynamics of the system. In our case, we expand this criterion that the horizon should be long enough to predict the states across the critical region, e.g., the corner or other geometry features where the scanning direction changes dramatically. This is to ensure that the MPC can handle extreme scenarios during the printing proactively by planning and optimizing the control input several steps ahead. The selection of the window size depends on the observability and noise of the system. As mentioned in [30], an auto-regressive representation for dynamical systems is required when the model is not fully observable. In our case, the chosen states are melt pool temperature and depth, which are the extracted representations of the melt pool dynamics, and might not be enough to accurately represent the system and treat it as a Markov Decision Process (MDP) by assuming that the system is fully observable. Therefore, an additional history is included to provide more information for better predictive accuracy. Further, since the past states (collected from the sensor feedback) are likely to be noisy due to the system variability and disturbance, the model prediction conditioned on a short but noisy history may result in greater prediction error as the critical features may be corrupted. In contrast, a larger window size will benefit the resilience of the prediction because the dense encoder of the TiDE model can filter the noise while still keeping critical information from the history, providing in-distribution prediction. Still, note that the optimization of $w$ and $p$ is beyond the scope of this work.

We generated a total of $N = 640,277$ time series segments, each with a length of 100, using $w = 50$ and $p = 50$. These were split into training/validation sets with a 9:1 ratio. The hyperparameters of the multi-variate model (predicting temperature and depth) and training setup are detailed in Table 2. We also trained a uni-variate TiDE for temperature prediction to compare with the PID controller with only 200 epochs. The TiDE model was trained using the Adam optimizer with a learning rate scheduler that decays the rate by 5% every two steps. Robustness and generality are key priorities during training. Since sensor noise and environmental uncertainties can corrupt past target data, the model must maintain predictive power even when noisy past targets and covariates are used. Additionally, smooth predictions of future states are essential for improving the MPC performance, even with noisy historical data. To achieve these improvements, regularization and dropout techniques were applied. For future work, combining experimental and simulation data using the co-teaching method [51] may further improve the model generality to noisy history. The validation of TiDE will be discussed in Section 6.1.

## 5. Model predictive control

### 5.1. Multi-step MPC formulation for DED with constraints

The multi-step MPC for melt pool temperature tracking and constraining melt pool depth at time $k$ can be formulated as follows:

$$\min_{\mathbf{u}^f = [u_k, \dots, u_{k+p-1}]} \sum_{i=1}^{p} \left[ \| \hat{x}_{temp,k+i} - r_{temp,k+i} \|_{\mathbf{Q}}^2 + \| \Delta u_{k+i-1} \|_{\mathbf{R}}^2 \right], \tag{8a}$$

$$s.t. \quad g_1(\mathbf{x}) : \hat{x}_{depth,k+i} \geq x_{depth}^{lb}, \quad \forall i \in \mathbb{N}_{[1,p-1]}, \tag{8b}$$

$$g_2(\mathbf{x}) : \hat{x}_{depth,k+i} \leq x_{depth}^{ub}, \quad \forall i \in \mathbb{N}_{[1,p-1]}, \tag{8c}$$

$$[\hat{x}_{temp}^{f+1}, \hat{x}_{depth}^{f+1}]^T = \mathrm{TiDE}(\mathbf{x}_{temp}^{p+1}, \mathbf{x}_{depth}^{p+1}, \mathbf{d}_x^{p:f},$$
$$\mathbf{d}_y^{p:f}, \mathbf{z}^{p:f}, \mathbf{u}^{p:f}) \tag{8d}$$

$$u_{k+i} \in \mathbb{U} := \{u \in \mathbb{R} \mid 504 \ W \leq u_i \leq 750 \ W\}, \tag{8e}$$

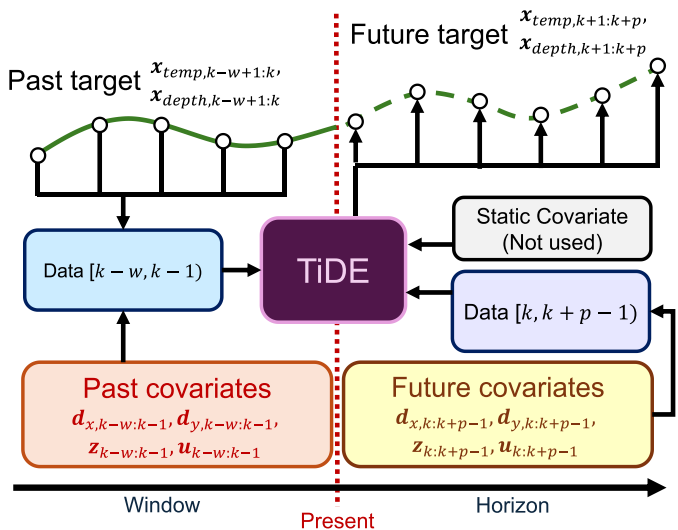

**Fig. 7.** Data structure of the input and output of TiDE, crafted for DED.

**Table 2**
Hyperparameters of training TiDE model.

| Details for TiDE model setup | | | | | | |
|---|---|---|---|---|---|---|
| # encoder layers | # decoder_layers | Decoder output dim. | Hidden size | Decoder hidden size | Dropout rate | Layer normalization |
| 1 | 1 | 16 | 128 | 32 | 0.2 | True |
| Details for TiDE model training | | | | | | |
| Learning rate | Regularization | Step_size | Rate decay | # epoch | Batch size | Shuffle data |
| 0.001 | 0.001 | 2 | 0.95 | 500 | 128 | True |

where $\Delta u_{k+i-1} = u_{k+i} - u_{k+i-1}$ represents the differences between two consecutive terms in the designed future laser power. To simplify the notation, we use the superscript $p$, $f$, and $p : f$ to denote the past $(k - w : k - 1)$, the future $(k : k + p - 1)$, and the past and future $(k - w : k + p - 1)$, respectively.

In practice, tuning the laser power is essential for ensuring the complete melting of the powder and maintaining melt pool temperature stability while minimizing the effects of overheating [11]. In this work, the purpose of implementing the MPC in DED is to provide a proactive control policy that mitigates defects by enforcing melt pool depth constraints when an arbitrary reference trajectory for melt pool temperature is provided. In DED, porosity is the most common and critical defect, directly affecting the mechanical properties of printed parts [1,52]. There are two main types of porosity: interlayer porosity, typically caused by low dilution, and intralayer porosity, resulting from high dilution. To mitigate these defects, it is recommended to maintain the melt pool depth within a dilution range of 10% to 30% [52]. While offline process optimization can generate melt pool temperature references that improve the properties of the fabricated material, it overlooks constraining the melt pool depth [7,11]. As a result, it potentially leads to the violation of the suggested dilution range during implementation, resulting in porosity defects. In such cases, the MPC can prioritize part quality over strict temperature tracking. Even when melting pool depth constraints are considered offline, the MPC can still act as a safeguard to ensure these constraints are consistently met. Lastly, although the sensors for in-situ melt pool depth measurements are almost unavailable at the current stage, we assume that the melt pool depth is observable via inference methods [43–45].

The MPC objective function includes the mismatch between the predicted future melt pool temperature and the reference trajectory, represented by the sum of square error, as well as the control effort, represented by the sum of $\Delta u_{k+i-1}$. The two types of loss are balanced by the weighting matrices $\mathbf{Q} = \mathbf{I}_p$ and $\mathbf{R} = 10\mathbf{I}_p$ in this work. The constraints involve maintaining the melt pool depth within bounds, specifically $x_{\text{depth}}^{lb} = 0.075$ mm and $x_{\text{depth}}^{ub} = 0.225$ mm. These constraints are enforced only after the first layer, as the melt pool depth from the GAMMA simulation remains below 10% dilution due to the substrate's boundary conditions during the first three layers. Additionally, the constraints are not considered near corners, i.e., $d_x \le 2$ mm∩$d_y \le 2$ mm, due to the unavoidable heat accumulation. The trained TiDE model is embedded into the MPC as the prediction model providing $\hat{x}^f$, and requires only one forward pass to generate the full prediction over the defined horizon.

### 5.2. Optimization setup

Several techniques have been implemented in this work to accelerate the solving process in real-time optimization. First, the gradient computation is handled using automatic differentiation through PyTorch's autograd, enabling efficient calculation of first-order derivatives without relying on numerical approximations [53]. For optimization, we employed the `l-bfgs` algorithm [54], implemented in the `PyTorch-minimize` package [55], which offers a balance between performance in large-scale optimization and computational efficiency, as it avoids the need for second-order derivative calculations (i.e.,

Hessians). Further, we employed a warm-start strategy in each MPC step to accelerate the solving process, using the optimal solution from the previous step as the initial guess for the current one.

Next, the constrained optimization problem in the MPC is reformulated as an unconstrained optimization problem using the augmented Lagrangian method, which transforms constraints into penalty [54]. This approach avoids the complexity of explicitly managing constraints (i.e., satisfying optimality conditions), and also enables the optimization to proceed in a smooth, continuous space, enhancing the solver's efficiency. Even so, implementing the augmented Lagrangian method will still increase the complexity of the objective function, and thus the optimizer might fail to terminate successfully. To maintain the feasibility of the MPC, when the solver terminates unsuccessfully, we resolve the problem using the default initial starting point instead of warm-start.

### 5.3. Execution of the MPC with GAMMA

Our pipeline for integrating the MPC with DED in GAMMA simulation is shown in Fig. 8. Due to the requirement of past covariate and target for TiDE, we first simulate GAMMA in an open-loop manner, then let the MPC take over the rest of the process using closed-loop control as the required past targets are collected. The MPC updates the control action (the first element of the optimal control sequence $\mathbf{u}^*$) every five GAMMA simulation timestep (i.e., 0.0355 s/iter for the MPC and 0.0071 s/iter. for GAMMA). In other words, the GAMMA will simulate the fabrication using the same control input throughout five simulation steps.

Since the step size of the laser nozzle toolpath does not match the element size, the extracted melt pool temperature and depth exhibit significant fluctuations and need to be filtered. These fluctuations are primarily periodic, arising from the mismatch between element size and scanning rate. To address this, we average the melt pool temperature and depth extracted from the GAMMA simulation $\mathbf{x}^{GAMMA}$ over the past 10 simulation steps as the measurement for the current MPC step $\mathbf{x}^{MPC}$, i.e., $\mathbf{x}_k^{MPC} = (\mathbf{x}_l^{GAMMA} + \mathbf{x}_{l-1}^{GAMMA} + \cdots + \mathbf{x}_{l-9}^{GAMMA})/10$, where $k$ denotes the MPC step and $l$ the GAMMA simulation step. This approach effectively reduces fluctuations in the extracted data from GAMMA.

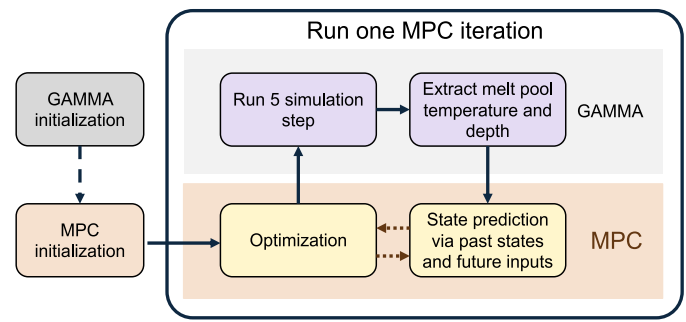

**Fig. 8.** Code pipeline for the execution of the MPC in GAMMA simulation.

## 5.4. Bench marking: PID controller

To evaluate the performance of the proposed MPC framework, we implemented a PID controller for melt pool temperature tracking as a benchmark, following [56]. Since the PID controller is primarily suited for single-input, single-output (SISO) systems and lacks intrinsic mechanisms to handle constraints explicitly, the comparison focuses solely on temperature reference tracking, with the melt pool depth constraint ignored for this benchmarking. The control input provided by the discrete-time PID controller can be obtained by:

$$u_k = u_{k-1} + K_p e_k + K_i e_{0:k} + K_d \frac{e_k - e_{k-1}}{\Delta t}, \tag{9}$$

where $e_k = r_k - x_{\text{temp},k}$ is the error between the reference and the current melt pool temperature, $e_{0:k} = e_{0:k-1} + e_k \Delta t$ is the discrete integral of error from $k = 0$, and $\Delta t$ is the MPC time step. The PID gains are optimized using Bayesian optimization [57] to minimize the mean square error between the trajectory and the reference. The PID controller operates with the same time step as the MPC and employs the same data smoothing method for data extraction.

## 6. Results

### 6.1. Model evaluation

In this section, we evaluate the TiDE model using the GAMMA simulation results with unseen laser power inputs as the ground truth for future targets. A local comparison between the TiDE predictions and the test set is illustrated in Fig. 9. To emphasize the model performance, two challenging scenarios during a single printing process – layer transitions and turning at corners – are highlighted in Fig. 9(a)(b) and (c)(d), respectively, where the melt pool temperature and depth show significant rises and drops. In Fig. 9(a), both the uni-variate and multi-variate TiDE models successfully capture the changes in melt pool temperature after the laser is turned off and then reactivated while it starts printing a new layer. Similarly, Fig. 9(b) demonstrates that the TiDE model accurately tracks the dynamics of the melt pool depth during this transition phase.

Furthermore, Fig. 9(c)(d) highlights how the TiDE models capture the temperature and depth dynamics when the laser nozzle turns at a corner, where heat tends to accumulate due to the change of laser speed and direction. The discrepancy between the predicted and actual temperatures at the corner is within the range of 5 to 15 K, as shown in Fig. 9(c). Notably, the TiDE models provide smoother predictions for temperature and depth compared to the more fluctuating ground truth values.

On a global scale, we assess the model's accuracy using the test set. The mean absolute percentage error (MAPE) and relative root mean square error (RRMSE) for melt pool temperature prediction are 1.29% and 0.054, respectively, for the uni-variate TiDE model, and 1.24% and 0.0515 for the multi-variate model. Additionally, for depth prediction, the multi-variate TiDE model achieves a MAPE of 4.25% and an RRMSE of 0.0441, indicating the high accuracy of the model.

Finally, we present the loss history for both uni-variate and multi-variate TiDE models in Fig. 9(e)(f). The validation loss consistently converges without signs of overfitting, despite the large fluctuations in the training loss. These validations confirm that the TiDE model is accurate and reliable for use in model predictive control (MPC) applications.

### 6.2. Melt pool temperature control using the MPC

We first demonstrate the implementation of the proposed MPC for tracking an arbitrary melt pool temperature reference, comparing its performance against a PID controller as a benchmark. Fig. 10(a) illustrates the complete temperature trajectory over 10 printed layers, comparing PID and the MPC using uni-variate TiDE prediction during the MPC. Data from the layer transition phases, where the laser is turned off, has been removed for clarity. Although TiDE predicts the entire horizon at each iteration, we only display the state prediction at the first step to simplify visualization. The MPC clearly produces a smoother trajectory than the PID controller, with TiDE predictions closely following the MPC trajectory, demonstrating the accuracy of the model.

This advantage of the MPC is further reflected in the applied laser input shown in Fig. 10(b). While both the MPC and the PID controllers apply similar trends in laser power, the MPC results in smoother inputs, reducing fluctuations. Notably, the MPC prevents the peak laser power observed at the beginning of each layer in the PID controller, thanks to its ability to anticipate the rise in melt pool temperature.

Three key scenarios from the printing process are highlighted: The MPC's takeover at the start (Fig. 10(c)), corner transitions on each layer (Fig. 10(d)), and the start of a new layer (Fig. 10(e)). In Fig. 10(c), the MPC demonstrates superior reference tracking compared to PID which exhibits more fluctuations. In Fig. 10(d), the MPC minimizes overshoot at corners, as it accounts for the learned dynamics, whereas PID produces larger tracking errors, resulting in greater overshoots. Similarly, Fig. 10(e) shows a more pronounced overshoot in the PID controller, while the MPC maintains a smoother trajectory, leveraging its predictive capabilities for better reference tracking.

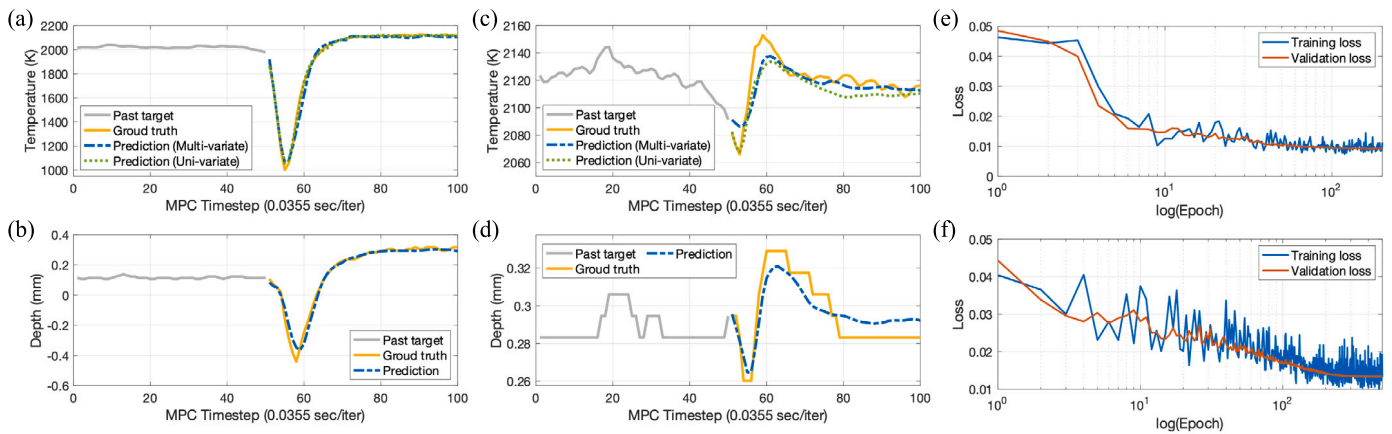

**Fig. 9.** Evaluation of TiDE model and training loss: (a) and (c) show the melt pool temperature comparison between the GAMMA simulation result as the ground truth and the predictions of uni-variate and multi-variate TiDE model, and (b) and (d) shows the melt pool depth comparison between GAMMA simulation and multi-variate TiDE model. The TiDE prediction is made in one-shot throughout the horizon $p = 50$. (e) and (f) are the loss for training uni-variate and multi-variate TiDE model.

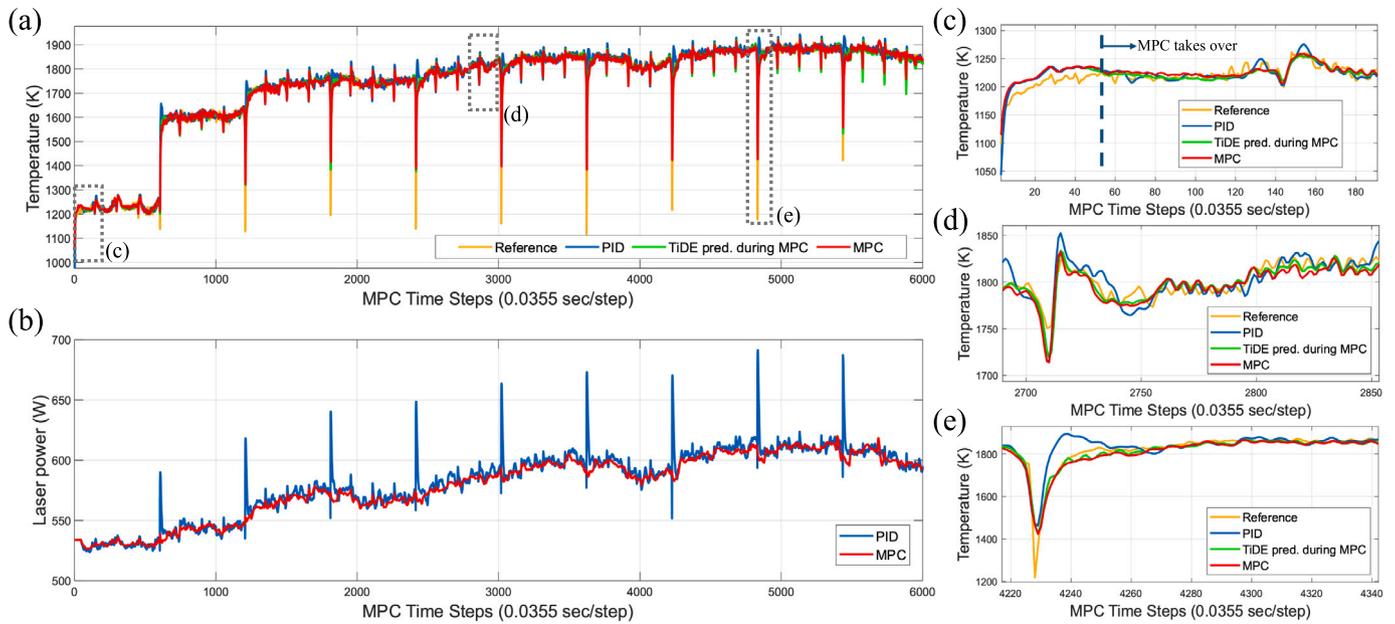

**Fig. 10.** Simulation result from GAMMA using the MPC and PID as controller. (a) Melt pool temperature reference, GAMMA simulation result using PID controller and the MPC, with the TiDE prediction used in the MPC. (b) Comparison of the controlled laser power using PID and the MPC, respectively. (c) Highlight of the beginning of the printing where the MPC starts activating. (d) Temperature trajectory at the corner. (e) Temperature trajectory at layer transition.

In all scenarios, TiDE model predictions closely align with the MPC trajectory, verifying accuracy. The $R^2$ values for MPC and PID trajectories, compared to the reference, are 0.9907 and 0.9827, respectively, indicating competitive performance. The developed MPC matches PID performance while significantly reducing overshoots and smoothing control inputs. However, the advantage of the MPC in DED extends beyond reference tracking to explicitly handle constraints in highly variable environments, as demonstrated in the next section.

### 6.3. Melt pool temperature control with melt pool depth as constraints

We further demonstrate the MPC's constraint-handling capability to prevent defects by keeping the melt pool depth within the 10%–30% dilution range. Fig. 11(a)–(c) compares the melt pool temperature, depth, and laser power trajectories, respectively, for the constrained and unconstrained MPC. The gray dashed boxes are the region for closer examination, detailed in Fig. 11(d)–(e).

In Fig. 11(a), the temperature profiles are nearly identical when the constraints are relaxed, with minor differences caused by GPU computation randomness. When constraints are enforced, the constrained MPC sacrifices certain performance of reference tracking to satisfy the constraints, leading to a significant deviation from the reference. Despite the fluctuations induced by constraint enforcement, the TiDE prediction accurately captures the system response. In Fig. 11(b), the constrained MPC effectively bounds the melt pool depth, with only minor violations occurring primarily at the start of a new layer, where some deviation is inevitable. In contrast, the unconstrained MPC maintains temperature tracking via the multi-variate TiDE model, but the melt pool depth exceeds 30% dilution after the fifth layer.

One consequence of constraint handling is the increased fluctuations in laser power, as shown in Fig. 11(c). Unlike the smooth trajectory of the unconstrained MPC, the constrained MPC continuously adjusts the laser power to balance constraint satisfaction and reference tracking. Additional fluctuations arise from occasional unsuccessful MPC solutions, where the system either reuses the previous iteration's solution or restarts with the default initial guess. This issue is exacerbated by the penalty method—if the MPC begins in an infeasible region, penalties can increase sharply, presenting a challenge in finding feasible solutions.

We also select two segments of the trajectories for detailed examinations. The first region, shown in Fig. 11(d), demonstrates the trajectories in layer 5, where we can observe how the MPC leverages tracking capability and constraint satisfaction. This segment also includes two corners where both the melt pool temperature and depth change dramatically due to the change in laser speed and direction. When the laser nozzle is entering the corner, although the penalty is ignored locally, the MPC decreases the laser power to prevent violating depth constraints. As the MPC anticipates that the margin exists between the current melt pool depth and the constraint, the laser power increases so that the tracking error can be reduced. In Fig. 11(e), we show the segment at the transition between layers 6 and 7. For the unconstrained MPC, the controlled melt pool temperature closely follows the reference well, while the melt pool depth often exceeds the upper bound. In contrast, for the constrained MPC, the melt pool depth is well maintained within the feasible region with only a few violations. Note that the melt pool depth in our simulation will always start from negative values as a new layer starts, where it is natural that the melt pool depth will be below 10% dilution. As shown in Fig. 11(e), the melt pool temperature is more sensitive to changes in laser power compared to the depth, with temperature exhibiting more significant variations under different inputs. This highlights the importance of the MPC in this multi-output system, where the optimal control sequence must account for a longer horizon to balance control objectives, especially when the dynamics of multiple responses are different.

### 6.4. Computational time

The histogram of solving times for both constrained and the unconstrained MPC, specifically during the solving process, is shown in Fig. 12, computed using an AMD Ryzen Threadripper PRO 3975WX 32-Cores CPU. For the unconstrained MPC, the mean solving time is 0.2575 s, with a maximum of 0.5437 s. In contrast, the solving time for the constrained MPC, i.e., the solving time using the warm start plus the backup initial guess if necessary, yields an average of 0.2775 s.

However, the distribution of the solving time has a long tail with 0.3% of the data greater than one second, showing that the augmented Lagrangian approach might introduce instability when the initial guess is infeasible and the penalty increases drastically. This issue can be potential solved by policy-based approaches such as explicit MPC [30,58] and ML-assisted MPC [59]. The former involves emulating the entire MPC-solving process using a policy learning model, and providing the optimal solution directly from the policy model, bypassing the solving process of MPC. The latter one focuses on improving the quality of warm start, rather than simply using the solution from the previous timestep. We observe the key issue that can fail the optimizer is when the reference trajectory changes significantly between consecutive steps, making the previous solution a poor initial guess. To address this, we can leverage the policy model (or explicit MPC) to provide a better warm start, ensuring that the warm start is closer to the optimum. This approach, as demonstrated in [59], shows that ML-assisted warm starts can significantly reduce the number of iterations required. Moreover, this method benefits from the initial guess from the policy model while still enforcing MPC constraints, ensuring safety guarantees.

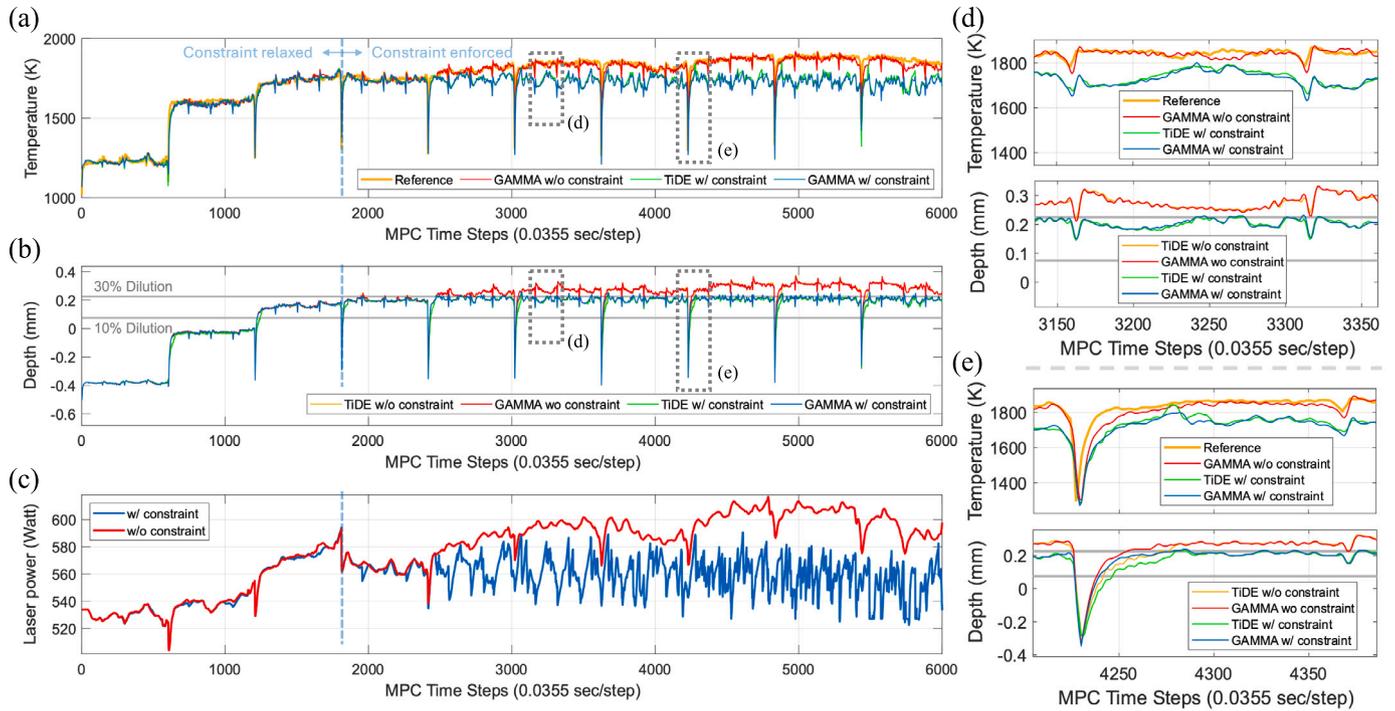

**Fig. 11.** Comparison between the enforcement of constraints on melt pool depth using GAMMA simulation. (a) Melt pool temperature comparison between the GAMMA result of unconstrained and the constrained MPC, with the TiDE prediction from the constrained MPC. (b) Melt pool depth comparison between the GAMMA result of the unconstrained and constrained MPC. (c) Applied laser power comparison for the constrained and unconstrained MPC. (d) Melt pool temperature and depth trajectory at the sixth layer. (e) Melt pool temperature and depth trajectory at the transition from the seventh to eighth layer.

### 6.5. Discussion

The verification using real-world experiments is planned in the long term, as it requires embedding the MPC controller with the existing machine, and building the interface connecting the sensors with the control units. Also, the implementation of a melt pool depth prediction module using online image processing will be required. In a well-controlled environment, we anticipate that the proposed MPC framework can perform as expected and will not be influenced by the environment noise for two reasons. First, because of the mismatch of laser scanning rate and element size, the uneven heat treatment time on each element induces severe fluctuations. Compared to the sensor data, the fluctuation that exists in the post-processed simulation data still exhibits greater variation than the noisy sensor data. Second, since TiDE uses the dense encoder to extract important features from the input data, it also serves as a noise filter to handle the noisy and corrupted data and exhibit noise-agnostic performance on prediction. Therefore, we believe that static environmental noise will not significantly affect the performance of the proposed MPC. However, the challenge of implementing the MPC lies in the mismatch between the model and the unknown while varying environmental conditions and material properties. Even though TiDE can be calibrated offline, an effective, data-efficient approach for model adaptation is required to address the unknown variability during the manufacturing process. As TiDE is a neural network (NN)-based model, it behaves similarly to other NNs and supports fine-tuning through approaches like few-shot learning [60,61].

We also expect that this approach is generalizable to other manufacturing systems. Since the simultaneous multi-step MPC is a variant of discrete-time MPC using a multi-step ahead predictor to replace recursive rollouts with the single-step ahead predictor, its formulation can be generalized to any type of dynamical system formulated in discrete state space systems. The examples in [33] show that a similar formulation can be applied to chemical engineering problems. We also use a discrete state space model as an illustration in our latest work [62]. For manufacturing systems, we expect that the proposed MPC can be applied to similar processes such as welding, as it also involves complex dynamics with temporal dependence on history and with a similar nature of metal deposition, liquidization, and solidification. Moreover, one special feature of our method compared to the existing multi-step MPC is that the TiDE model is capable of accommodating the pre-defined parameters, e.g. geometrical information and processing condition (material) as the model input using the covariates, making it suitable for manufacturing systems where the geometry and the path can be pre-determined, increasing the predictive accuracy of TiDE. The necessary steps for adapting the proposed framework to other systems will be identifying important features of the geometry, such as feature engineering, to represent the local geometry using the covariates. Also, we suggest that the fine-tuning of the hyperparameters $(w, p, Q, R)$ may be important to balance between proactive and conservative control strategies. In the broad application, we also expect that the proposed multi-step MPC approach can be used in scheduling problems where the MPC can be integrated with more existing time-series applications to perform a broad Digital Twin application in decision-making.

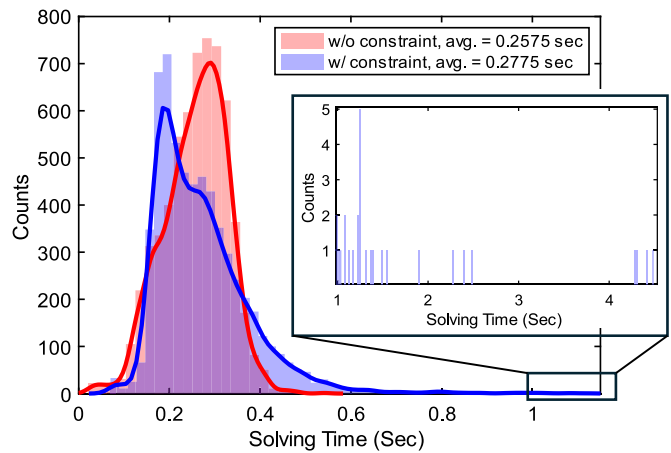

**Fig. 12.** Histogram of solving time for the constrained and unconstrained MPC. The highlighted window shows the counts of instances with solving time greater than 1 s.

While this work presents a promising approach, several assumptions were made that may limit its general applicability. First, we assume that the computational time for solving the MPC and TiDE evaluations can be neglected, thus potential delays were not considered. In reality, the solving time of the MPC is not always bounded, making it difficult to determine a consistent control frequency. This issue may be solved by using neural network-based optimization (NNBO) methods [28] or explicit MPC [58] to emulate the MPC-solving process, allowing a single neural network to provide control policies in a single forward pass. Additionally, in our framework, we fixed the scanning rate to simplify the processes of making the prediction and solving MPC. However, introducing the scanning rate as a dynamic control variable could potentially enhance the flexibility of the process control. Moreover, the TiDE model we developed is tailored to our specific target geometry, limiting its applicability to other geometries. Lastly, this work serves as a proof of concept using simulation data and has yet to be validated through experimental data, which could provide further insights into its practical effectiveness.

## 7. Closure

In this work, we introduce a simultaneous multi-step MPC framework using time-series DNN as an embodiment of real-time decision-making for Digital Twins in autonomous manufacturing. This framework utilizes a data-driven time-series model, TiDE, to predict future states required for the MPC in one shot, then implements gradient-based optimization for solving the MPC. Although we focus on DED as an example in this work, the demonstrated nonlinear system identification with TiDE and the multi-step MPC framework can be seamlessly generalized for other manufacturing systems. While using DED as the case study, our method leverages the strengths of TiDE, such as its ability to handle both past and future covariates and fast prediction, making it well-suited for the dynamic nature of DED and the MPC controllers, respectively. Through rigorous validation using a single-track multi-layer square, we demonstrate the accuracy and reliability of the TiDE model in predicting melt pool features. The results show that the proposed MPC effectively handles melt pool depth constraint, which is a challenging task for the PID controller, while yielding competitive performance as the PID controllers in tracking melt pool temperature. This work not only highlights the need for MPC to improve the quality of manufacturing processes but also opens new applications for integrating ML with MPC.

In the future, we will leverage the learned quantile [62] for uncertainty quantification to develop a robust MPC framework to improve guarantee constraint satisfaction. We will also explore geometric-agnostic model representations, such as the Koopman operator [31] or neural operators [63], alongside compatible MPC frameworks to enhance generality and knowledge transfer. Last but not least, efficient model updating techniques will be developed and integrated with our data-driven models, enabling real-time updates and decision-making to fulfill the full potential of a Digital Twin system.

## CRediT authorship contribution statement

**Yi-Ping Chen:** Conceptualization, Data curation, Formal analysis, Methodology, Project administration, Validation, Visualization, Writing – original draft, Writing – review & editing. **Vispi Karkaria:** Conceptualization, Formal analysis, Methodology, Validation, Writing – review & editing. **Ying-Kuan Tsai:** Conceptualization, Formal analysis, Methodology, Writing – review & editing. **Faith Rolark:** Data curation, Formal analysis, Software, Validation, Writing – original draft, Writing – review & editing. **Daniel Quispe:** Software, Writing – review & editing. **Robert X. Gao:** Conceptualization, Funding acquisition, Investigation, Supervision, Writing – review & editing. **Jian Cao:** Conceptualization, Funding acquisition, Investigation, Supervision, Writing – review & editing. **Wei Chen:** Conceptualization, Funding acquisition, Investigation, Methodology, Resources, Supervision, Writing – review & editing.

## Declaration of competing interest

The authors declare that they have no known competing financial interests or personal relationships that could have appeared to influence the work reported in this paper.

## Acknowledgments

We appreciate the grant support from the NSF HAMMER-ERC (Engineering Research Center for Hybrid Autonomous Manufacturing, Moving from Evolution to Revolution) under Award Number EEC-2133630, and the NSF MADE-PUBLIC Future Manufacturing Research Grant Program under Award Number CMMI-2037026. Yi-Ping Chen also appreciates the Taiwan-Northwestern Doctoral Scholarship funded by the Ministry of Education in Taiwan, and the fellowship support from the Predictive Science and Engineering Design (PSED) cluster at Northwestern University.